\documentclass{ieeeojies}

\usepackage{cite}
\usepackage{amsmath,amssymb,amsfonts}
\pdfoutput=1
\usepackage{graphicx}
\usepackage{textcomp}

\def\BibTeX{{\rm B\kern-.05em{\sc i\kern-.025em b}\kern-.08em
    T\kern-.1667em\lower.7ex\hbox{E}\kern-.125emX}}

\usepackage{graphicx}
\usepackage{mathtools}

\usepackage[none]{hyphenat}
\usepackage{textcomp}
\usepackage{array}
\usepackage{kantlipsum}
\usepackage{flexisym}
\usepackage{stackengine}
\usepackage{booktabs}
\usepackage{lipsum}
\usepackage{multirow}
\usepackage{subcaption}
\usepackage{breqn}

\usepackage{algorithm}

\usepackage{mathrsfs}
\usepackage[noend]{algpseudocode}
\usepackage{algpseudocode}

\usepackage[dvipsnames, table, xcdraw]{xcolor}

\begin{document}
\title{\centering \huge \color{black} Knowledge Distillation and Enhanced Subdomain Adaptation Using Graph Convolutional Network for Resource-Constrained Fault Diagnosis}
\author{\uppercase{Mohammadreza Kavianpour}\authorrefmark{1},
\uppercase{Parisa Kavianpour\authorrefmark{2}, \uppercase{Amin Ramezani}\authorrefmark{1,3}, and Mohammad TH Beheshti}.\authorrefmark{1}}
\address[1]{Department of Electrical and Computer Engineering, Tarbiat Modares University, Iran. (e-mail: kavianpour, ramezani, mbehesht@modares.ac.ir)}
\address[2]{Faculty of Engineering and Technology, University of Mazandaran, Iran (e-mail: p.kavianpour08@umail.umz.ac.ir)}
\address[3]{Department of Medicine, Baylor College of Medicine, USA (email: amin.ramezani@bcm.edu)}

\markboth
{Kavianpour \headeretal: Knowledge Distillation and Enhanced Subdomain Adaptation Using GCN for Resource-Constrained Bearing Fault Diagnosis}
{Kavianpour \headeretal: Knowledge Distillation and Enhanced Subdomain Adaptation Using GCN for Resource-Constrained Bearing Fault Diagnosis}

\corresp{Corresponding author: Mohammadreza Kavianpour (e-mail: kavianpour@modares.ac.ir).}

\begin{abstract}
Bearing fault diagnosis under varying working conditions faces challenges, including lack of labeled data, distribution discrepancies, and resource constraints. To address these issues, we propose a progressive knowledge distillation framework that transfers knowledge from a complex teacher model, utilizing a Graph Convolutional Network (GCN) with Auto-Regressive Moving Average (ARMA) filters, to a compact and efficient student model. To mitigate distribution discrepancies and labeling uncertainty, we introduce Enhanced Local Maximum Mean square Discrepancy (ELMMSD), which leverages mean and variance statistics in the Reproducing Kernel Hilbert Space (RKHS) and incorporates a priori probability distributions between labels. This approach increases the distance between clustering centers, bridges subdomain gaps, and enhances subdomain alignment reliability. Experimental results on benchmark datasets (CWRU and JNU) demonstrate that the proposed method achieves superior diagnostic accuracy while significantly reducing computational costs. Comprehensive ablation studies validate each component's effectiveness, highlighting the approach's robustness and adaptability across diverse working conditions.
\end{abstract}


\titlepgskip=-15pt

\maketitle

\section{Introduction}

\subsection{Bearing Fault Diagnosis}

Rotating machinery plays a crucial role in various industrial applications, with bearings being among the most failure-prone components. Timely and accurate detection of bearing faults is essential to prevent significant financial losses and operational downtime \cite{qian2025federated,liu2025mpnet}. Recent advancements in Deep Learning (DL) methods, particularly Convolutional Neural Networks (CNNs) \cite{kavianpour2021earthquake, kavianpour2023cnn} and Graph Convolutional Networks (GCNs) \cite{liu2025few,wang2023attention}, have shown considerable promise in enhancing fault diagnosis accuracy. However, the practical implementation of these methods in industrial settings faces three primary challenges: computational complexity, domain shift, and limited access to high-quality labeled data.

The computational complexity of DL models presents significant hurdles for deployment on resource-constrained devices \cite{liu2025lmsff,kavianpour2021intelligent,hu2025lightweight, kavianpour2022deep}. Striking an optimal balance between model accuracy and efficiency is vital for real-world applications. Techniques such as quantization \cite{dantas2024comprehensive}, low-rank approximation \cite{huang2024iteratively}, network pruning \cite{ruan2023light}, and Knowledge Distillation (KD) \cite{guo2025multilevel} have been developed to mitigate these issues. Among these, KD has emerged as a compelling approach that facilitates the transfer of knowledge from high-performing teacher models to lightweight student models. Nonetheless, traditional KD methods often encounter limitations, including a capacity gap between teacher and student models, which can hinder the effectiveness of knowledge transfer in complex scenarios.

Another critical challenge is domain shift, arising from discrepancies between training data (source domain) and operational data (target domain). Domain adaptation (DA) methods, including Maximum Mean Discrepancy (MMD) \cite{yu2024m,9737184} and Correlation Alignment (CORAL) \cite{shao2024adaptive,kavianpour2022intelligent}, are commonly employed to bridge this gap. However, these methods often face limitations when applied to high-dimensional vibration signals, as they primarily focus on aligning mean statistics or covariances, failing to capture the complete distributional characteristics of the data \cite{qian2023maximum,kavianpour2024physics}. Moreover, their performance heavily depends on high-quality labels \cite{fu2025gsscl, yang2024towards} and neglects consideration for aligning subdomains with identical classes \cite{qian2023cross}. To address this latter limitation, Subdomain Adaptation (SDA) techniques, such as Local Maximum Mean Discrepancy (LMMD) \cite{chen2025rigid}, have been proposed to align both marginal and conditional distributions by focusing on localized variations. However, the challenges of addressing high-dimensional data complexities and reliance on high-quality labels remain unresolved.

\subsection{Related Works}

Significant progress has been made in unsupervised bearing fault diagnosis under varying operating conditions. Research efforts have primarily focused on minimizing domain gaps through DA, SDA, or a combination of both. While earlier studies predominantly employed CNN-based methods, recent work has explored the advantages of GCNs in capturing structural features. Below, we review key developments in DA, SDA, and GCN-based approaches, followed by an analysis of KD methods and their limitations.

\subsubsection*{Domain Adaptation and Subdomain Adaptation}

Qian et al. \cite{qian2023deep} combined CORAL and MMD to address domain gaps arising from changing working conditions, introducing an I-SoftMax function to enhance classification. Similarly, Jiang et al. \cite{jiang2023deep} utilized LMMD and CORAL within a CNN framework to minimize domain discrepancies. Liu et al. \cite{liu2024cross} proposed a subdomain adaptation technique integrating local correlation alignment with statistical and geometric feature extraction to capture fine-grained details. Ding et al. \cite{ding2023deep} applied LMMD to address feature and label shifts under varying conditions, while Li et al. \cite{li2024fault} employed multi-kernel MMD for distribution alignment in rotary machinery fault diagnosis. Liang et al. \cite{liang2023unsupervised} incorporated deformable convolutional layers and LLMD into a residual network for fault diagnosis under time-varying speeds. Although these methods help reduce domain gaps, they often overlook practical computational constraints and fail to leverage the structural learning capabilities of GCNs. Additionally, by focusing primarily on the mean of data distributions, they neglect variance, which limits their robustness. Their sensitivity to noisy labels further constrains generalization under uncertain conditions.

\subsubsection*{GCN-Based Methods}

GCN models have demonstrated potential in bearing fault diagnosis by effectively extracting structural features. Li et al. \cite{li2020multireceptive} introduced a multi-receptive field GCN to capture features from neighboring nodes. Sun et al. \cite{sun2021multi} proposed a multi-scale cluster graph convolutional network incorporating residual blocks and auto-encoder-based graph generation layers to extract weak signal features. Yu et al. \cite{yu2021fault} employed a ChebyNet-based GCN to process wavelet packet-transformed signals, focusing on single-neighbor connections for feature learning. Xu et al. \cite{xu2023graph} designed a graph-guided collaborative convolutional network with a reasoning fusion module to analyze correlations between multi-source signals. Darvishi et al. \cite{darvishi2023deep} applied recurrent GCNs for complex feature extraction, while Ghorvei et al. \cite{ghorvei2023spatial} combined CNNs with topology-adaptive GCNs using DANN and LMMD for reducing distribution discrepancy. Zhang et al. \cite{zhang2024cross} integrated polynomial-based GCN filters with multi-kernel MMD for domain alignment. Chen et al. \cite{chen2024deep} combined CNN and GCN with hybrid LLMD and CDAN for improved adaptation. However, Standard graph filters like Chebyshev and polynomial types struggle with inflexibility, over-smoothing, and computational costs, especially with higher degrees. Spectral filters, while powerful, are computationally demanding, non-local, and sensitive to graph structure variations. These challenges underscore the need for robust and transferable GCN filters capable of effectively capturing geometric features.

\subsubsection*{Knowledge Distillation-Based Methods}

Knowledge distillation has been widely utilized to transfer insights from complex teacher models to smaller, more efficient student models. Chen et al. \cite{chen2025multi} demonstrated this by distilling knowledge from a multi-scale graph pyramid attention network (MsGPAT) to a simplified student model, although without addressing domain adaptation. Gue et al. \cite{guo2024lightweight} combined KD with uniform quantization to compress large models while using DANN to minimize domain gaps in the teacher model. Westfechtel et al. \cite{westfechtel2023combining} introduced adversarial domain adaptation for teacher models while employing KD loss to guide student models. Ji et al. \cite{ji2022neural} proposed a KD method incorporating parameter quantization tailored for resource-constrained environments but overlooked varying operational conditions. Lu et al. \cite{lu2024lightweight} integrated KD with multi-kernel MMD for efficient fault diagnosis, while Pan et al. \cite{pan2024global} developed a joint KD strategy for domain-shared and domain-specific knowledge transfer. Despite these advancements, current KD methods do not fully address subdomain adaptation or the challenges posed by noisy labels. Traditional KD approaches often compress teacher models before adaptation, which can diminish their generalization capacity. Conversely, adapting teacher models prior to compression is challenging, particularly due to the dependence of unsupervised KD on labeled target domain data.

\subsection{Solution and Contribution}

To address the aforementioned challenges, this study proposes a \textbf{K}nowledge distill\textbf{A}tion of graph con\textbf{V}olutional neural networks with subdoma\textbf{I}n adaptation called KAVI for resource-constrained fault diagnosis. The framework integrates progressive knowledge distillation with a subdomain adaptation technique, leveraging multi-kernel enhanced local maximum mean square discrepancy (ELMMSD). The proposed ELMMSD metric extends conventional methods by considering both the mean and variance of data distributions within the Reproducing Kernel Hilbert Space (RKHS), enabling more precise alignment of complex domain distributions while reducing computational complexity. The KAVI method employs progressive KD to gradually align the student model, using the teacher's knowledge to learn domain-invariant features. Furthermore, a label smoothing (LS) strategy is employed to introduce prior probabilities that modify hard labels into a softer distribution, enhancing decision boundaries and mitigating the impact of noisy labels. This approach helps reduce the model's overconfidence and improves its generalization performance.
\par The framework utilizes GCNs enhanced with auto-regressive moving average (ARMA) filters in the teacher model to extract geometric structural features, improving subdomain adaptation by capturing flexible frequency responses, enhancing transferability across graphs with different structures, and reducing noise sensitivity. A compact CNN architecture serves as the lightweight student model, optimized through progressive knowledge distillation to incrementally align with the target domain. This approach effectively balances accuracy, efficiency, and adaptability, addressing the computational and domain shift challenges in fault diagnosis.

The main contributions of this study are:

\begin{itemize}
    \item \textbf{Enhanced Local Maximum Mean Square Discrepancy (ELMMSD)}: A novel metric that captures both the mean and variance of data distributions in RKHS for precise domain alignment, combined with label smoothing to enhance robustness against noisy labels and improves classification accuracy.
    \item \textbf{Progressive Knowledge Distillation Framework:} An approach that reduces the student model size by 99.67\% (from 0.92 MB to 0.028 MB) while maintaining a negligible accuracy loss of 2\%, enabling deployment on resource-constrained devices.
    \item \textbf{ GCNs with ARMA Filters:} Leveraging GCNs with ARMA filters to capture geometric structural features, enhancing subdomain adaptation and robustness to domain shifts.
    \item \textbf{Unified Fault Diagnosis Solution:} A comprehensive framework integrating ELMMSD, knowledge distillation, and GCNs to address computational complexity, domain shift, and noisy label challenges in bearing fault diagnosis.
\end{itemize}

This framework demonstrates significant potential for practical application in resource-constrained industrial environments, offering a robust and efficient solution to bearing fault diagnosis.

\section {Preliminaries}
\subsection{Problem formulation}

Consider the source domain ${\mathscr{D}_s} = \left\{ {\left( {{x^{s}_{i}},{y^{s}_{i}}} \right)} \right\}_{i = 1}^{{n_s}}$, containing \(n_s\) labeled samples, and the target domain ${\mathscr{D}_t} = \left\{ {\left( {{x^{t}_{j}}} \right)} \right\}_{j = 1}^{n_t}$, comprising \(n_t\) unlabeled samples. Here, \(x^{s}_{i}\) and \(x^{t}_{j}\) refer to the feature representations in the source and target domains, respectively, while \(y^{s}_{i}\) denotes the labels in the source domain. The datasets \(\mathscr{D}_s\) and \(\mathscr{D}_t\)  have identical $n_c$ classes and are sampled from distinct but related distributions, \(p_s\) and \(p_t\). This distributional divergence, where \(p_s \neq p_t\), presents a domain shift challenge that impedes the straightforward transfer of models trained on \(\mathscr{D}_s\) to \(\mathscr{D}_t\).

\subsection{Graph Convolutional Network with ARMA filter}
Consider a graph $\mathcal{G}$ with $N$ nodes, represented as $\mathcal{G} = (\mathcal{V}, \mathscr{E}, A)$, where $\mathcal{V} = \{\mathcal{V}_{1}, \mathcal{V}_{2}, ..., \mathcal{V}_{N}\}$ denotes the set of nodes, $\mathscr{E}$ is the set of edges, and $A \in \mathbb{R}^{N \times N}$ is the adjacency matrix defining connectivity. An entry $A_{i,j} = 1$ indicates a link between nodes $\mathcal{V}_{i}$ and $\mathcal{V}_{j}$, and $A_{i,j} = 0$ otherwise. The Laplacian matrix, $L = D - A$, is derived from the degree matrix $D \in \mathbb{R}^{N \times N}$, where $D_{i,i} = \sum_j A_{i,j}$. The symmetric Laplacian ${L}^{sym}$, used frequently in spectral analysis, is expressed as:
\begin{equation}
    L^{sym} = I_N - D^{-1/2} A D^{-1/2} = U \Lambda U^T,
\end{equation}
where $I_N$ is the identity matrix, $\Lambda = \text{diag}(\lambda_1, \lambda_2, ..., \lambda_N)$ holds the eigenvalues, and $U$ is the matrix of eigenvectors. A graph filter modifies signals on the graph using the spectral decomposition of $L^{sym}$, yielding the filtered signal $\bar{X}$:
\begin{equation}
    \bar{X} = \sum_{i=1}^N h(\lambda_i) u_i u_i^T X = U \, \text{diag}[h(\lambda_1), ..., h(\lambda_N)] U^T X,
\end{equation}
where $h(\lambda_i)$ represents the filter's frequency response for each eigenvalue $\lambda_i$.

However, spectral filters have limitations, such as high computational cost and lack of localization, affecting all nodes regardless of proximity. To overcome these, the ARMA filter is introduced. The $K$-order ARMA filter with coefficients of $a$ and $b$ is defined as:
\begin{equation}
    h_{ARMA_K}(\lambda) = \frac{\sum_{k=0}^{K-1} b_k \lambda^k}{1 + \sum_{k=1}^K a_k \lambda^k},
\end{equation}
which translates in the node domain to:
\begin{equation}
    \bar{X} = \left(I + \sum_{k=1}^K a_k L^k\right)^{-1} \left(\sum_{k=0}^{K-1} b_k L^k\right) X.
\end{equation}

To reduce computational complexity, a first-order ARMA$_1$ filter is implemented using a recursive approximation:
\begin{equation}
    \bar{X}^{(t+1)} = p F \bar{X}^{(t)} + q X,
\end{equation}
where $F = \frac{1}{2}(\lambda_{\max} - \lambda_{\min}) I - L$, with $p$ and $q$ as coefficients, and $\lambda_{\max}$, $\lambda_{\min}$ denoting the Laplacian's maximum and minimum eigenvalues. The filter’s convergence leads to:
\begin{equation}
    \bar{X} = \lim_{t \to \infty} \left[(p F)^t \bar{X}^{(0)} + q \sum_{j=0}^t (p F)^j X \right].
\end{equation}

This recursive ARMA$_1$ formulation effectively localizes filtering and adapts well to diverse graph architectures. Higher-order ARMA filters are constructed by combining multiple ARMA$_1$ filters:
\begin{equation}
    \bar{X} = \sum_{f=1}^F h_{ARMA_K}(\gamma_f) u_f u_f^T X,
\end{equation}
where $u_f$ is eigenvector, $\gamma_f = \frac{1}{2}(\lambda_{\max} - \lambda_{\min}) - \lambda_{f}$, and
\begin{equation}
    h_{ARMA_K}(\gamma_f) = \sum_{k=1}^K \frac{q_k}{1 - p_k \gamma_f}.
\end{equation}

Adjusting the coefficients $a_k$ allows transformation between ARMA and polynomial filters, offering flexibility in the filter design \cite{kavianpour2022class,bianchi2021graph}.

\subsection{maximum mean square discrepancy}\label{MMSD}
MMSD is an extension of MMD designed to measure distributional differences between two domains. In MMSD, the kernel function \( k(x, .) \) maps samples into a RKHS, where the discrepancies are computed using the tensor product \( k(x, .) \otimes k(x, .) \). This tensor product approach retains the functional properties of MMD while providing enhanced sensitivity to higher-order statistics and reducing computational complexity.

The MMSD formulation begins by defining a RKHS 
\[
\mathcal{H} = \text{span}\{k(x, .) \mid x \in X\}.
\]
The operational mechanics leverage the kernel's role in mapping samples into an infinite-dimensional space, enabling effective measurement of distribution gaps. For two domains, \(\mathscr{D}_s\) and \(\mathscr{D}_t\), the MMSD is defined as:
\begin{equation}
    \begin{split}
        & 
        {MMSD}[\mathcal{H} \otimes \mathcal{H}, \mathscr{D}_s, \mathscr{D}_t] = \\
        &
        \sup_{\|h\|_{\mathcal{H}_1 \otimes \mathcal{H}_2} \leq 1} 
    \Big| \langle h, \mathbb{E}_{p_s}[k(x^s, .) \otimes k(x^s, .)] \rangle_{\mathcal{H}} - \\
        & 
    \langle h, \mathbb{E}_{p_t}[k(., x^t) \otimes k(., x^t)] \rangle_{\mathcal{H}} \Big|.
    \end{split}
\end{equation}

where $<.,.>$ is the inner product, $\mathbb{E}$ is expected value, $\mathcal{H}$ indicates a Hilbert space, and $h \in \mathcal{H}$. The tensor product structure ensures that the MMSD accounts for higher-order relationships between distributions, with the kernel's tensor product used as a substitute for traditional inner products. This substitution preserves the methodology’s robustness while enhancing computational efficiency.

The empirical formulation of MMSD introduces a biased statistic for practical scenarios:

\begin{dmath}
\label{MMSD3}
    $$MMSD_b^2[\mathcal{H} \otimes \mathcal{H},{\mathscr{D}_s},{\mathscr{D}_t}] =  \frac{1}{{n_s^2}}\sum\limits_{i = 1}^{{n_s}} {\sum\limits_{j = 1}^{{n_s}} {{k^2}(x_i^s,x_j^s)}  + } \frac{1}{{n_t^2}}\sum\limits_{i = 1}^{{n_t}} {\sum\limits_{j = 1}^{{n_t}} {{k^2}(x_i^t,x_j^t)} }  - \frac{2}{{{n_s}{n_t}}}\sum\limits_{i = 1}^{{n_s}} {\sum\limits_{j = 1}^{{n_t}} {{k^2}(x_i^s,x_j^t)} } $$
\end{dmath}

\section{Proposed Method}

\begin{figure*}[th]

    \centering
    \centering
    \includegraphics[width=0.8\textwidth]{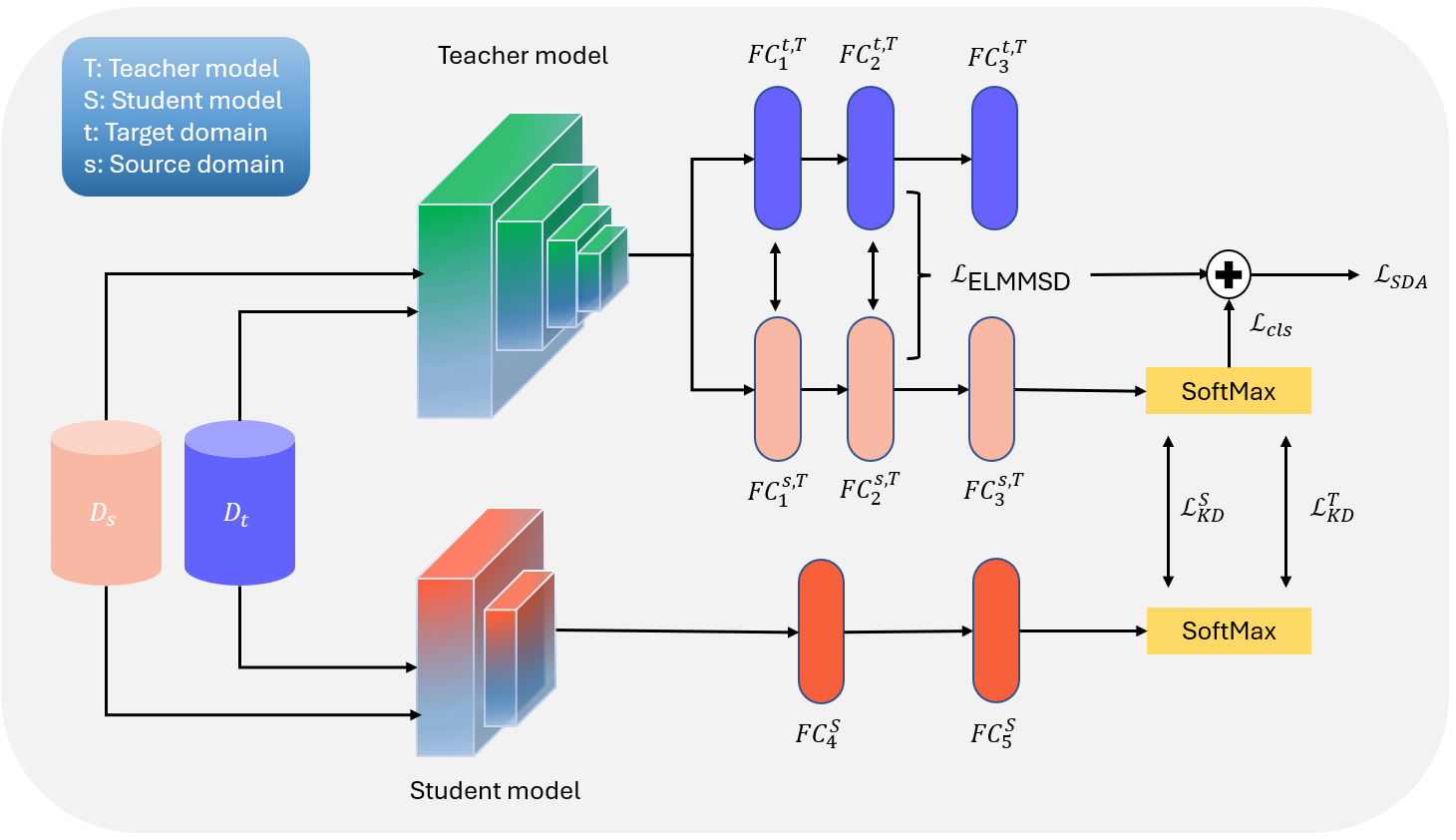}
    \caption{schematic diagram of proposed  method}
    \label{proposedd}
 
    \end{figure*}

This section details the structure of the proposed KAVI method, whose overall framework is depicted in Figure~\ref{proposedd}. The approach comprises three main modules: (1) Teacher and Student Models, (2) Subdomain Adaptation Module, and (3) Progressive Knowledge Distillation Module.

\subsection{Module 1: Teacher and Student Models}

In the first module, vibrational data from the source and target domains are fed into Teacher and Student models. The Teacher model employs a complex architecture based on a Graph Convolutional Neural Network (GCN), while the Student model adopts a simpler Convolutional Neural Network (CNN) to reduce computational complexity and facilitate deployment. The structure of Teacher and Student models is provided in Table \ref{structure}

\subsubsection{Teacher Model Architecture}

% \begin{figure*}[th]

%     \centering
%     \centering
%     \includegraphics[width=\textwidth]{1.JPG}
%     \caption{schematic diagram of proposed  method}
%     \label{proposedd}
 
%     \end{figure*}

The Teacher model begins by constructing \textit{instance graphs} from each mini-batch using a Graph Generation Layer (GGL) \cite{li2021domain}. GGL generates input graphs for the ARMA-based graph convolutional layers. The adjacency matrix \(A\) is computed by normalizing the product of the feature matrix \(\mathscr{X}\) and its transpose:

\begin{equation}
    A = \text{normalize}(\mathscr{X} \mathscr{X}^T).
\end{equation}

To reduce computational cost and ensure sparsity, the \(Top-\mathcal{K}\) largest values of \(A\) are selected row-wise using a sparse indexing matrix \(\mathcal{K}(\cdot)\):

\begin{equation}
    \tilde{A} = Top-\mathcal{K}(A),
\end{equation}

where \(\mathcal{K}\) is set to 2. The resulting graph is then processed through three ARMA\(_1\) convolutional layers to extract structural information. The ARMA\(_1\) convolution operation is defined as:

\begin{equation}
    \tilde{\mathscr{X}}^{(t+1)} = \operatorname{ReLU}\left(\tilde{F}\tilde{\mathscr{X}}^{(t)}W + \tilde{\mathscr{X}}V\right),
\end{equation}

where \(W\) and \(V\) are learnable parameters, \(\tilde{F}\) represents the modified Laplacian, and \(\lambda_{\text{max}} = 2\), \(\lambda_{\text{min}} = 0\). This simplifies computation while allowing \(W\) and \(V\) to offset minor errors caused by the assumption.

The output of the \(l^{\text{th}}\) ARMA\(_1\) layer is computed recursively as:

\begin{equation}
    \text{ARMA}_1^l = H(\tilde{\mathscr{X}}^{(l-1)}),
\end{equation}

where \(H\) denotes the ARMA\(_1\) graph convolution operation from the previous layer. Finally, ARMA\(_K\) filters are applied recursively across three stacked layers for enhanced feature extraction. After each GCN layer, ReLU activation and batch normalization are employed. Features from the final ARMA\(_K\) layer are passed through three fully connected layers to diagnose faults and minimize structural distribution differences between domains. 

\subsubsection{Student Model Architecture}

The Student model is designed for low-cost deployment, utilizing a lightweight one-dimensional CNN with two layers. This design ensures fewer parameters and reduced computational overhead while maintaining effectiveness.

\subsection{Subdomain Adaptation Module}

The Subdomain Adaptation Module is introduced to reduce discrepancies in subdomain distributions within the Teacher model. The varying operational conditions of rotating machinery result in differences in the marginal and conditional distributions across these domains. To address this, the proposed ELMMSD method integrates the strengths of LMMD, LS, and MMSD, providing a more nuanced and effective alignment of global domain distributions while ensuring precise subdomain matching, particularly in complex and heterogeneous scenarios. Consequently, this approach aligns both marginal and conditional distributions of the source and target domains, leading to a robust and accurate knowledge transfer. 

In the following subsection, we define the embedding label smoothing method in the new ELMMSD method using the relationships defined for MMSD in subsection \ref{MMSD} and LMMD in \cite{ghorvei2021unsupervised}.

\subsubsection{Label Smoothing}

Traditional LMMD effectively aligns subspaces across domains but demands high-quality class labels. When source domain labels are noisy or uncertain, the reliance on cross-entropy loss—designed to optimize correct labels while neglecting incorrect ones—undermines its ability to compute discrepancy distances accurately. This limitation weakens network generalization, increasing the risks of domain drift and negative transfer. To address these challenges, we propose incorporating Label Smoothing, a technique proven effective in Computer Vision, to enhance reliability in subdomain alignment.

Label smoothing modifies deterministic labels by introducing a probabilistic representation, balancing between original hard labels and smoothed distributions. This approach prevents overconfidence in label predictions and facilitates the model's capacity to capture subtle distinctions between subcategories. By softening labels, ELMMSD effectively minimizes ambiguities between closely related fault types, resulting in tighter clustering for similar faults and greater separation among dissimilar classes in the target domain. Compared to conventional LMMD, ELMMSD improves the decision boundary for subdomain adaptation by leveraging these refined label representations.

Mathematically, the initial prediction logits \( z_i^s \) for each category in labeled source domain are converted to probabilities via Softmax:
\begin{equation}
    p(c \mid x_i^s) = \frac{\exp(z_i^s)}{\sum_{j=1}^{n_c} \exp(z_j^s)}.
\end{equation}

The Cross-Entropy (CE) loss function is defined as:
\begin{equation}
    h_s(p, q) = -\sum_{c=1}^{n_c} q(c \mid x_i^s) \log(p(c \mid x_i^s)),
\end{equation}
where \( q \) denotes the true label distribution. Using LS, the smoothed label \( S(y_i) \) is computed as:
\begin{equation}
    S(y_i) = (1 - \epsilon)y_i + \frac{\epsilon}{n_c},
\end{equation}
where \( \epsilon \) is the smoothing coefficient, and \( y_i \) is the one-hot encoded label. Integrating \( S(y_i) \) into the CE loss yields the smoothed classification loss:
\begin{equation}\label{l_cls}
    \mathscr{L}_{\text{cls}} = -\sum_{c=1}^C \left[ (1 - \epsilon)q(c \mid x_i^s) + \frac{\epsilon}{n_c} \right] \log(p(c \mid x_i^s)),
\end{equation}
% with gradients computed as:
% \begin{equation}
%     \nabla l_{\text{cls}} = p(c \mid x_i^s) - (1 - \epsilon)q(c \mid x_i^s) - \frac{\epsilon}{n_c}.
% \end{equation}

When employing subdomain discrepancy distance, the network's fully connected (FC) layer typically outputs the sample data, with source domain labels represented in one-hot encoding. Their smoothed representation is formulated as:

\begin{equation} \label{z_hat}
    \hat{z}_i^s = (1 - \epsilon)y_i^{\text{s}} + \frac{\epsilon}{C}
\end{equation}

\subsubsection{Definition of Enhanced Local Maximum Mean square Discrepancy}
Based on the definitions of LMMD loss, label smoothing, and MMSD distance, we extend this framework by introducing multi-layer multi-kernel ELMMSD within two fully connected layers, FC1 and FC2. The alignment is formulated as follows:

\begin{dmath}
\label{f3}
{{\hat d}_{z_1}} = \frac{1}{{{n_c}}}\sum\limits_{c = 1}^{{n_c}} {\sum\limits_{i = 1}^{{n_s}} {\sum\limits_{j = 1}^{{n_s}} {\omega _i^{sc}\omega _j^{sc}{k^2}\left( {\hat{z}_i^{1s},\hat{z}_j^{1s}} \right)} } }  
+ \frac{1}{{{n_c}}}\sum\limits_{c = 1}^{{n_c}} {\sum\limits_{i = 1}^{{n_t}} {\sum\limits_{j = 1}^{{n_t}} {\omega _i^{tc}\omega _j^{tc}{k^2}\left( {z_i^{1t},z_j^{1t}} \right)} } }  
- \frac{2}{{{n_c}}}\sum\limits_{c = 1}^{{n_c}} {\sum\limits_{i = 1}^{{n_s}} {\sum\limits_{j = 1}^{{n_t}} {\omega _i^{sc}\omega _j^{tc}{k^2}\left( {\hat{z}_i^{1s},z_j^{1t}} \right)} } }.
\end{dmath}

\begin{dmath}
\label{f4}
{{\hat d}_{z_2}} = \frac{1}{{{n_c}}}\sum\limits_{c = 1}^{{n_c}} {\sum\limits_{i = 1}^{{n_s}} {\sum\limits_{j = 1}^{{n_s}} {\omega _i^{sc}\omega _j^{sc}{k^2}\left( {\hat{z}_i^{2s},\hat{z}_j^{2s}} \right)} } }  
+ \frac{1}{{{n_c}}}\sum\limits_{c = 1}^{{n_c}} {\sum\limits_{i = 1}^{{n_t}} {\sum\limits_{j = 1}^{{n_t}} {\omega _i^{tc}\omega _j^{tc}{k^2}\left( {z_i^{2t},z_j^{2t}} \right)} } }  
- \frac{2}{{{n_c}}}\sum\limits_{c = 1}^{{n_c}} {\sum\limits_{i = 1}^{{n_s}} {\sum\limits_{j = 1}^{{n_t}} {\omega _i^{sc}\omega _j^{tc}{k^2}\left( {\hat{z}_i^{2s},z_j^{2t}} \right)} } }.
\end{dmath}

Here, $z_1$ and $z_2$ are the outputs of the FC1 and FC2 layers and   $\hat{z}_1$ and $\hat{z}_2$ represent the outputs of these layers using smooth labeling calculated by Eq. \ref{z_hat}, respectively. ${\omega_i^{sc}}$ and ${\omega_j^{tc}}$ denote the weights for the $c^{\text{th}}$ class in the source and target domains. Moreover, ELMMSD replaces inner products with tensor products of kernel functions, retaining the operational properties of LMMD while reducing computational complexity. This ensures that MMSD is computationally efficient despite its higher-order representation.

Given that kernel selection significantly influences mean embedding matching, multiple kernels with varying bandwidths are employed to capture both low-order and high-order moments of the features, thereby minimizing domain discrepancies. Accordingly, the kernel function in Equations \eqref{f3} and \eqref{f4} is defined as:

\begin{equation}
k \triangleq \sum\limits_{u = 1}^U \mu_u k_u,
\end{equation}

The total objective function of subdomain adaptation derived from the multi-layer multi-kernel ELMMSD and cross-entropy loss is expressed as:
\begin{dmath}\label{SDA}
        {\mathscr{L}_{SDA}} = {\mathscr{L}_{CLS}}+ \lambda_{SDA} ({{\hat d}_{z_1}} + {{\hat d}_{z_2}})
\end{dmath}

where in this study, we use a dynamic trade-off factor to improve model training performance and the transferability of extracted features. This factor is defined as follows:
\begin{equation} \label{lambda_sda}
\lambda_{SDA} = -\frac{4}{\sqrt{\frac{e}{n_{e}+1}}+1}+ 4
\end{equation}

where $e$ and $n_{e}$ are epoch and maximum epochs, respectively. Initially, we set $\lambda_{SDA}$ to 0 to facilitate the learning of fundamental fault characteristics. As the training progresses and the number of epochs increases, $\lambda_{SDA}$ gradually rises, allowing for more effective learning of transferable features based on the fault characteristics learned from the source domain, activating the ELMMSD loss.

\subsection{Progressive Knowledge Distillation Module}

Following the adaptation of subdomains in the Teacher model, the knowledge derived from unlabeled target data must be effectively transferred to the Student model. To achieve this, the target knowledge distillation loss function ${\mathscr{L}_{KD}^T}$ is defined as:

\begin{equation}
\label{ltkd1}
    \begin{aligned}
        {\mathscr{L}_{KD}^T} = {\mathscr{L}_{KL}}({Q_s}(\mathscr{D}_t,\tau ),{Q_t}(\mathscr{D}_t,\tau ))
    \end{aligned}
\end{equation}

where ${Q_s}$ and ${Q_t}$ denote the outputs of the Student and Teacher models, respectively, for unlabeled inputs from the target domain. A temperature-based softmax activation function is employed to smooth the outputs, defined as:

\begin{equation}
\label{ltkd}
    \begin{aligned}
        {Q_i} = \frac{{\exp ({\raise0.7ex\hbox{${{z_i}}$} \!\mathord{\left/
 {\vphantom {{{z_i}} \tau }}\right.\kern-\nulldelimiterspace}
\!\lower0.7ex\hbox{$\tau $}})}}{{\sum\nolimits_j {\exp ({\raise0.7ex\hbox{${{z_j}}$} \!\mathord{\left/
 {\vphantom {{{z_j}} \tau }}\right.\kern-\nulldelimiterspace}
\!\lower0.7ex\hbox{$\tau $}})} }}
    \end{aligned}
\end{equation} 

Here, $z_i$ represents the logits of the \( i^{\text{th}} \) class in the output layer, and $\tau$ is the temperature parameter. Larger values of $\tau$ produce softer outputs. The loss function ${\mathscr{L}_{KL}}$ in Eq. \ref{ltkd1} is computed using the Kullback–Leibler divergence. 

Additionally, to ensure that the Student model effectively learns features from labeled source data, the loss function ${\mathscr{L}_{KD}^S}$ is defined as:

\begin{dmath}
\label{lskd}
   {\mathscr{L}_{KD}^S} = {\mathscr{L}_{KL}}({Q_s}(\mathscr{D}_s,\tau ),{Q_t}(\mathscr{D}_s,\tau )) + \lambda_{CLS} {\mathscr{L}_{CLS}}
\end{dmath}

where $\lambda_{CLS}$ is a trade-off coefficient balancing the contributions of $\mathscr{L}_{KL}$ and $\mathscr{L}_{CLS}$. 

The overall objective function of the proposed model integrates the loss functions from Eq. \ref{SDA}, Eq. \ref{ltkd}, and Eq. \ref{lskd}, expressed as:

\begin{equation}\label{total_loss}
    \begin{aligned}
        {\mathscr{L}_{total}} = (1 - \lambda_e){\mathscr{L}_{SDA}} + \lambda_e ({\mathscr{L}_{KD}^T} + {\mathscr{L}_{KD}^S})
    \end{aligned}
\end{equation}

The parameter $\lambda_e$ acts as a trade-off factor between the SDA and KD modules. Early in the training process, due to the substantial distribution discrepancy between domains, limited target knowledge can be effectively transferred. Thus, $\lambda_e$ is initialized to a low value and increases exponentially with training progress. If $\lambda_e$ is bounded in the range $[\alpha_1, \alpha_2]$, its value at epoch $e$ is updated as:

\begin{equation}\label{lambda}
    {\lambda _e} = {\alpha _1}.\exp \left(\frac{e}{{n_e}} \cdot \log \left(\frac{{\alpha _2}}{{\alpha _1}}\right)\right)
\end{equation}

The workflow of the proposed method is illustrated in Algorithm \ref{step2}.

\begin{algorithm}[h]
\caption{Distilling knowledge from the Teacher model being adapted to the Student model}\label{step2}
\begin{algorithmic}  
    \State\textbf{Input}\textit{ source domain $\mathscr{D}_s$, target domain $\mathscr{D}_t$, GCN-based Teacher model ($T$), CNN-based Student model ($S$), number of epochs ($n_e$)} 
    \State\textit{Generate graph with GGL}
    \State\textit{Initialize parameters of Teacher and Student models}

    \For { \textit{$e=1,...,n_e$}}
        \For{\textit{$x_s\in \mathscr{D}_s$ and $x_t\in \mathscr{D}_t  $}}
            \State\textit{Extract local-global discriminative features}
            \State\textit{Generate target pseudo labels}
            \State\textit{Calculate smoothed labels $\hat{z}_i^s$ by Eq. \ref{z_hat}}
            \State\textit{Calculate smoothed CE loss ($\mathscr{L}_{\text{cls}}$) by Eq. \ref{l_cls}}
            \State\textit{Calculate SDA loss ($\mathscr{L}_{SDA}$) by Eq. \ref{SDA}}
            \State\textit{Calculate KD losses ${\mathscr{L}_{KD}^T}$ and ${\mathscr{L}_{KD}^S}$ in models by Eq. \ref{ltkd1} and Eq. \ref{lskd}}

		 %  \State\textit{Optimize $(1-\lambda){\mathscr{L}_{DA}}$ for model $T$ using $x_s$ and $x_t$ } 
			% \State \textit{Optimize $\beta{\mathscr{L}_{KD}^S}$ for models $S$ and $T$ using $x_s$}
   %          \State\textit{Optimize $\beta{\mathscr{L}_{KD}^T}$ for models $S$ and $T$ using $x_t$}

            \State\textit{Calculate total loss by Eq. \ref{total_loss}}
            \State\textit{update weights of the KAVI model using backpropagation algorithm}
            \State\textit{Update $\lambda_{SDA}$ using Eq. \ref{lambda_sda}}
            \State\textit{Update $\lambda$ using Eq. \ref{lambda}}

        \EndFor
        \State\textbf{end for}

    \EndFor
    \State\textbf{end for}

	\State \textbf{Output:} \textit{a strong lightweight Student model}
\end{algorithmic}
\end{algorithm}

\section{Experiments}
\subsection{Dataset Description}

This study evaluates the proposed KAVI method using two benchmark datasets: the Case Western Reserve University (CWRU) dataset \cite{lou2004bearing} and the JiangNan University (JNU) dataset \cite{zhao2021applications}.

\subsubsection{CWRU Dataset}

The CWRU dataset includes data from ten bearing health states: a healthy state and three fault conditions—outer ring fault (ORF), inner ring fault (IRF), and ball fault (BF). Each fault type is represented at three severity levels (0.007, 0.014, and 0.021 inches). Vibration signals were captured using a drive-end sensor operating at a 12 kHz sampling frequency under four load conditions: 0 hp ($A1$), 1 hp ($A2$), 2 hp ($A3$), and 3 hp ($A4$). This setup allows for the definition of 12 distinct transfer learning tasks. Each task consists of ten classes, with 1,000 samples per class. Each sample contains 1,024 data points, extracted using an overlapping segmentation technique to ensure consistent sample lengths.

\subsubsection{JNU Dataset}

The JNU dataset was generated by inducing faults using wire-cutting machines and recording vibration data at a 50 kHz sampling rate for 30 seconds. Similar to the CWRU dataset, it encompasses four distinct health states. Data acquisition was performed under three rotational speeds: 600 rpm ($J1$), 800 rpm ($J2$), and 1,000 rpm ($J3$). This configuration supports six domain adaptation scenarios, where knowledge transfer occurs between speed-based domains. For example, the transfer task \(J1 \to J2\) represents adapting a model trained on data from 600 rpm (source domain) to data from 800 rpm (target domain).

\subsection{Implementation Details}

The architecture of the Student and Teacher models employed in this study is outlined in Table \ref{structure}. Training is conducted with a batch size of 128, and the optimization process utilizes stochastic gradient descent. The proposed method is trained over 400 epochs with an initial learning rate of 0.001. Key hyperparameters include a temperature parameter ($\tau$) set to 20 and a trade-off parameter $\lambda_e$ that starts at 0.1 and increases exponentially to 0.9. The coefficient $\lambda_{CLS}$ is fixed at 0.8 to balance the loss components. Datasets are divided into 70\% for training, 15\% for validation, and 15\% for testing. A third-order ARMA filter is employed to extract structural features. For ELMMSD, multi-Gaussian kernels with bandwidths of \{0.001, 0.01, 1, 10, 100\} are applied. Hyperparameter tuning is performed using a grid search to optimize performance metrics. Each experiment is repeated five times to ensure reproducibility, with final results reported as the average across trials. All experiments are implemented using the PyTorch framework.

% \usepackage{multirow}
% \usepackage{graphicx}
\begin{table}[]
\caption{The detail of proposed method's structure}
\label{structure}
\resizebox{\columnwidth}{!}{%
\begin{tabular}{ccc}
\hline
Model                          & Layers                       & Number of neurons or size/stride/number of kernels \\ \hline
\multirow{7}{*}{Teacher model} & GGL                          & -                                                  \\
                               & ARMA$^1$, BN, ReLU           & 128                                                \\
                               & ARMA$^2$, BN, ReLU           & 128                                                \\
                               & ARMA$^3$, BN, ReLU           & 128                                                \\
                               & FC1                          & 256                                                \\
                               & FC2                          & 128                                                \\
                               & FC3                          & Number of health states                            \\ \hline
\multirow{4}{*}{Student model} & Conv1, BN, ReLU, Max-pooling & 3/2/16                                             \\
                               & Conv2, BN, ReLU, GAP         & 3/2/32                                             \\
                               & FC4                          & 128                                                \\
                               & FC5                          & Number of health states                            \\ \hline
\end{tabular}%
}
\end{table}

\subsection{Experiments and Comparison Methods}

To evaluate the KAVI approach for bearing fault diagnosis under varying operational conditions, we conducted a series of experiments and ablation studies. These studies aim to validate the effectiveness of the proposed components and their contributions to the overall performance. Specifically, we focus on the following objectives:

\begin{itemize}
    \item Demonstrating the efficacy of the novel ELMMSD distance as an SDA technique.
    \item Assessing the impact of the proposed progressive knowledge distillation in transferring knowledge from a complex Teacher model to a lightweight Student model with minimal accuracy degradation.
    \item Justifying the use of the ARMA filter in the Teacher model’s GCN compared to state-of-the-art filters.
    \item Highlighting the benefits of the proposed smoothing label strategy in subdomain adaptation.
\end{itemize}

To systematically compare the proposed method, we categorize the experiments into three groups:

\paragraph{Category 1: KD and SDA Configurations}
This category investigates various configurations and sequences of knowledge distillation and SDA to evaluate their individual and combined effects on model performance. The configurations are as follows:

\begin{itemize}
    \item \textbf{SDA followed by KD (SDA → KD):} The Teacher model is first adapted to the target subdomain using SDA, followed by KD to transfer the adapted knowledge to the Student model. This configuration may limit the Student model's direct generalization to the target subdomain as SDA does not involve the Student model during adaptation.
    \item \textbf{KD followed by SDA (KD → SDA):} KD is initially applied to train the Student model using labeled source domain data, followed by SDA to adapt the Student model to the target subdomain. This sequential approach provides a solid foundational understanding of the source domain, potentially enhancing alignment with the target subdomain.
    \item \textbf{Direct SDA on the Student Model:} SDA is applied directly to the Student model without prior KD. This configuration evaluates the Student model's ability to independently adapt to the target subdomain, though it lacks structured guidance from the Teacher model.
\end{itemize}

\paragraph{Category 2: Effect of DA and SDA Techniques}
This category replaces the proposed ELMMSD with established DA and SDA techniques, such as DANN \cite{ganin2016domain}, LMMD \cite{yuan2024intelligent}, and MMSD \cite{qian2023maximum}, while using hard labels for comparison. These experiments aim to highlight the effectiveness of the LS factor and square kernels in ELMMSD for feature adaptation across domains compared to conventional methods. The baselines of both models in these techniques are the same as those of the proposed method.

\paragraph{Category 3: Effect of ARMA GCN}
To assess the contribution of the ARMA filter in the Teacher model's GCN, we substitute it with alternative backbones, including CNN, Graph attention network (GAT) \cite{velivckovic2017graph}, multi-receptive field graph convolutional network (MRFGCN) \cite{li2021domain}, and topology adaptive graph convolutional network (TAGCN) \cite{ghorvei2023spatial}. Additionally, the CNN baseline is constructed using three convolutional layers inspired by the Student model's architecture. These experiments provide insights into the performance of the ARMA GCN relative to state-of-the-art GCNs and traditional CNNs.

Each experiment is designed to isolate and analyze the contributions of individual components and configurations, providing a comprehensive evaluation of the proposed method under diverse conditions.

\section{result}
\subsection{Analysis of results}
Table \ref{config_KD} presents the accuracy achieved by different configurations of SDA and KD for randomly selected tasks on the CWRU and JNU datasets. The proposed KAVI method demonstrates superior performance across all tasks, outperforming the other configurations. The results indicate that the "SDA only" configuration performs better than the "SDA → KD" method. This outcome can be attributed to the absence of labels in the target dataset, wherein the "SDA → KD" configuration, relying solely on the distillation loss is insufficient; the cross-entropy loss is crucial to supervise adaptation effectively. 
The "KD → SDA" configuration, leveraging the benefits of both KD and SDA, achieves higher accuracy by providing a strong foundation for subsequent adaptation. However, KAVI further enhances this approach by progressively transferring knowledge from the Teacher to the Student model, leading to significant performance gains. KAVI's ability to bridge the performance gap between the Teacher and Student models without compromising accuracy is a key advantage. Moreover, its generalizability and model-agnostic nature make it applicable to a wide range of SDA and KD techniques, offering flexibility and adaptability in various fault diagnosis scenarios.

\par Table~\ref{acc_cwru} and Table~\ref{acc_jnu} present the accuracy of Teacher and Student models for different SDA and DA methods across all transfer tasks for the CWRU and JNU datasets, respectively. MMSD outperforms LMMD and DANN due to its ability to incorporate both mean and variance information, making it more robust and comprehensive in representing distribution discrepancies. The use of square kernels in MMSD enables the representation of second-order statistics, which are crucial for tasks involving complex and subtle domain shifts. The proposed KAVI method, leveraging the advantages of ELMMSD, further improves performance by incorporating smoothing labels and aligning both marginal and conditional distributions effectively.

\par
\begin{figure}[ht]
    \centering
    \includegraphics[width=\columnwidth]{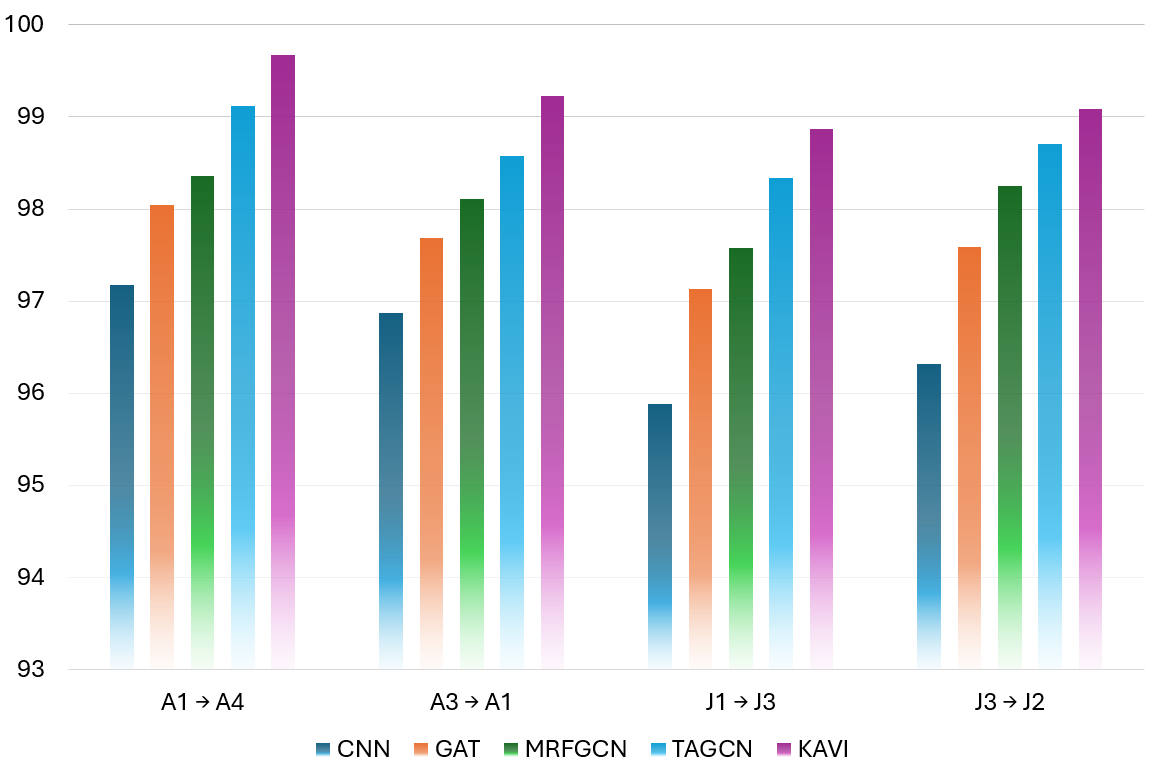}
    \caption{Comparison accuracy of some tasks on diverse baseline for Teacher model}
    \label{graph_result}
\end{figure}

 The diagnosis accuracy of various methods for Category 3 is presented in Fig. \ref{graph_result} for randomly selected transfer tasks. Among the evaluated approaches, the proposed KAVI method, integrating ARMA filters, consistently demonstrates superior performance, achieving the highest accuracy across all tasks. For instance, in the transfer task A1 → A4, KAVI achieves an accuracy of 99.67\%, surpassing TAGCN, the second-best performer, by 0.55\%. Similarly, in J3 → J2, KAVI achieves 99.09\%, reflecting a notable improvement of 0.38\% over TAGCN. These results highlight the robustness, transferability, and adaptability of ARMA filters in addressing complex graph-based relationships. Furthermore, all GCN-based methods outperform the CNN-based Teacher model, underscoring the effectiveness of GCN-based feature learning for fault diagnosis applications.

\begin{table}[t]
\caption{The accuracy of the Student model on diverse configurations for SDA and KD}
\label{config_KD}
\resizebox{\columnwidth}{!}{%
\begin{tabular}{ccccccc}
\hline
Methods & A1 $\to$ A2 & A2 $\to$ A4 & A4 $\to$ A1 & J2 $\to$ J1 & J2 $\to$ J3 & J3 $\to$ J1 \\ \hline
SDA → KD & 67.87 & 65.37 & 66.63 & 61.98 & 67.77 & 64.95 \\
KD → SDA & 95.17 & 94.78 & 94.83 & 93.77 & 94.58 & 94.42 \\
SDA only & 94.31 & 94.02 & 93.98 & 93.36 & 93.47 & 93.39 \\
KAVI (Student)    & 97.53 & 97.04 & 97.13 & 96.02 & 96.59 & 95.69 \\ \hline
\end{tabular}%
}
\end{table}

% \usepackage{multirow}
% \usepackage{graphicx}
\begin{table}[t]
\caption{The comparison of diagnosis accuracy of diverse methods on Teacher and Student networks for the CWRU dataset}
\label{acc_cwru}
\resizebox{\columnwidth}{!}{%
\begin{tabular}{cccccllll}
\hline
\multirow{2}{*}{Tasks} & \multicolumn{4}{c}{Teacher}   & \multicolumn{4}{c}{Student}   \\ \cline{2-9} 
 & KAVI & MMSD & LMMD & DANN & \multicolumn{1}{c}{KAVI} & \multicolumn{1}{c}{MMSD} & \multicolumn{1}{c}{LMMD} & \multicolumn{1}{c}{DANN} \\ \hline
A1 $\to$ A2            & 99.53 & 98.83 & 97.85 & 97.22 & 97.53 & 96.33 & 94.85 & 93.92 \\
A1 $\to$ A3            & 99.82 & 99.14 & 97.79 & 97.15 & 97.76 & 96.58 & 94.8  & 93.85 \\
A1 $\to$ A4            & 99.67 & 99.24 & 98.04 & 97.67 & 97.62 & 96.67 & 95.03 & 94.34 \\
A2 $\to$ A1            & 99.87 & 99.06 & 97.77 & 97.14 & 97.78 & 96.52 & 94.78 & 93.84 \\
A2 $\to$ A3            & 99.74 & 98.97 & 97.42 & 97.27 & 97.67 & 96.45 & 94.45 & 93.96 \\
A2 $\to$ A4            & 99.08 & 98.16 & 96.87 & 96.36 & 97.04 & 95.45 & 93.92 & 93.09 \\
A3 $\to$ A1            & 99.23 & 98.24 & 97.13 & 96.88 & 97.16 & 95.78 & 94.16 & 93.58 \\
A3 $\to$ A2            & 99.83 & 98.65 & 97.33 & 96.92 & 97.79 & 96.12 & 94.34 & 93.62 \\
A3 $\to$ A4            & 99.73 & 98.94 & 97.58 & 96.86 & 97.66 & 96.37 & 94.57 & 93.56 \\
A4 $\to$ A1            & 99.18 & 97.72 & 97.14 & 96.65 & 97.13 & 95.34 & 94.17 & 93.36 \\
A4 $\to$ A2            & 98.90 & 97.98 & 97.03 & 96.72 & 96.9  & 95.57 & 94.07 & 93.42 \\
A4 $\to$ A3            & 99.74 & 98.11 & 97.24 & 96.97 & 97.67 & 95.68 & 94.26 & 93.66 \\ \hline
\end{tabular}%
}
\end{table}

\begin{table}[t]
\caption{The comparison of diagnosis accuracy of diverse methods on Teacher and Student models for the JNU dataset}
\label{acc_jnu}
\resizebox{\columnwidth}{!}{%
\begin{tabular}{cccccllll}
\hline
\multirow{2}{*}{Tasks} & \multicolumn{4}{c}{Teacher}   & \multicolumn{4}{c}{Student}   \\ \cline{2-9} 
 & KAVI & MMSD & LMMD & DANN & \multicolumn{1}{c}{KAVI} & \multicolumn{1}{c}{MMSD} & \multicolumn{1}{c}{LMMD} & \multicolumn{1}{c}{DANN} \\ \hline
J1 $\to$ J2            & 98.93 & 97.51 & 96.76 & 96.33 & 96.68 & 94.92 & 93.34 & 93.23 \\
J1 $\to$ J3            & 98.87 & 98.56 & 96.05 & 95.92 & 96.23 & 94.75 & 92.68 & 92.71 \\
J2 $\to$ J1            & 98.46 & 97.31 & 95.93 & 95.66 & 96.02 & 95.36 & 93.43 & 92.41 \\
J2 $\to$ J3            & 99.17 & 98.43 & 96.61 & 96.15 & 96.59 & 94.87 & 94.58 & 93.87 \\
J3 $\to$ J1            & 98.75 & 98.37 & 95.88 & 95.34 & 95.69 & 94.72 & 94.1  & 92.62 \\
J3 $\to$ J2            & 99.09 & 98.65 & 96.34 & 95.93 & 96.56 & 95.65 & 94.54 & 94.17 \\ \hline
\end{tabular}%
}
\end{table}

\par
\begin{figure}[htbp]
    \centering
    \begin{subfigure}{0.45\textwidth}
        \centering
        \includegraphics[width=\linewidth]{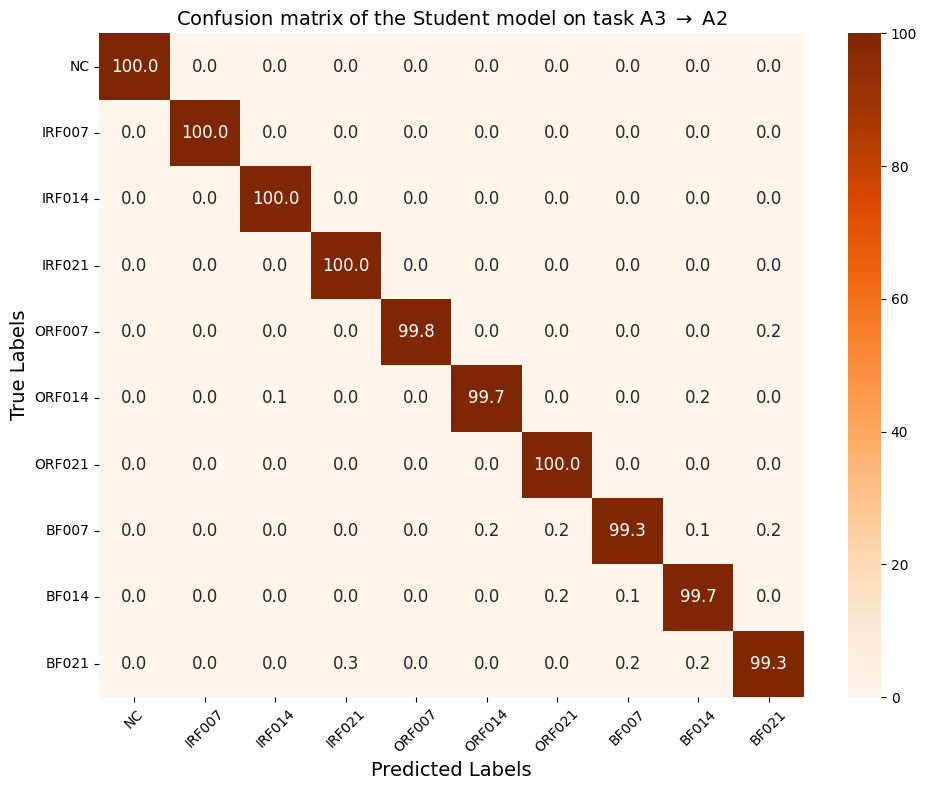}
        \caption{Student model}
        \label{fig:first_image}
    \end{subfigure}
    \hfill
    \begin{subfigure}{0.45\textwidth}
        \centering
        \includegraphics[width=\linewidth]{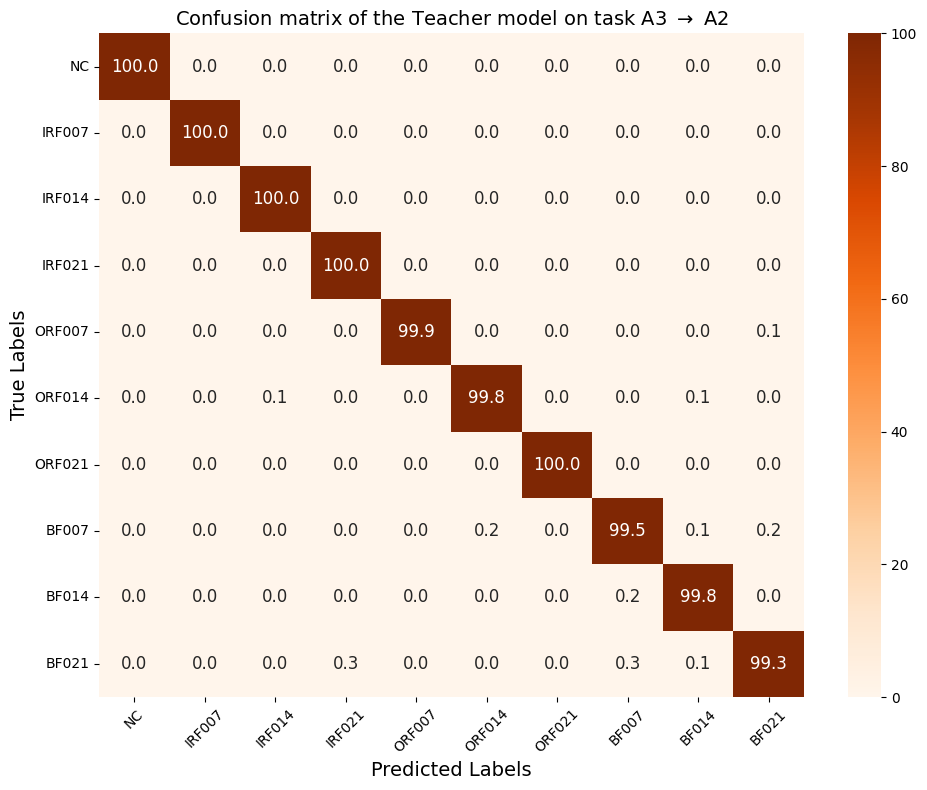}
        \caption{Teacher model}
        \label{fig:second_image}
    \end{subfigure}
    \caption{Confusion matrix of the KAVI method for task A3$\to$A2}
    \label{confusion_matrix}
\end{figure}

 These findings underscore the robustness and effectiveness of KAVI, demonstrating its potential to advance the state-of-the-art in bearing fault diagnosis under challenging operational conditions. In addition, Fig. \ref{confusion_matrix} exhibits the confusion matrix of the proposed method for task A3$\to$A2, which can be used to evaluate the KAVI method's performance in more detail for both Student and Teacher models. Fig. \ref{confusion_matrix} illustrates that the proposed method correctly identified all classes, with the lowest accuracy being 99.3\% for class BF021.

\subsection{Impact of ARMA Filter Node Configuration on Computational Cost and Accuracy}

To assess the impact of the number of nodes in the ARMA filter, we analyzed the model size, FLOPs, and accuracy of both the Teacher and Student models across different configurations with 32, 64, 128, and 256 nodes. The results, presented in Table \ref{model_size}, pertain to task A1 $\to$ A3. The findings reveal that using 128 nodes achieves the highest accuracy for both the Teacher (99.82\%) and Student (98.77\%) models. Notably, increasing the number of nodes to 256 results in a significant increase in FLOPs (126.16M) and model size (1.97MB), approximately 2.13 times larger than the configuration with 128 nodes, without improving the accuracy. This highlights the critical importance of carefully selecting hyperparameters to balance accuracy and computational efficiency. Also, it depicts that a bigger teacher model with 256 nodes may not aim to enhance performance because the learning bottleneck is now on the Student model.

Additionally, the FLOPs and model size for other tasks within the CWRU dataset have identical values to those reported in Table \ref{model_size}. When selecting 128 nodes for the ARMA filter, the Student model achieves a substantial reduction in computational cost, with FLOPs and model size reduced to 32.83M and 0.028MB, respectively, compared to the Teacher model’s 59.05M FLOPs and 0.92MB size. Despite this reduction, the Student model maintains a high accuracy of 97.76\%, demonstrating the effectiveness of the proposed method in achieving unsupervised fault diagnosis with minimal computational overhead.

% \usepackage{graphicx}
\begin{table}[]
\caption{Impact of ARMA Filter Nodes on Computational Cost and Accuracy for Task A1 $\to$ A3}
\label{model_size}
\resizebox{\columnwidth}{!}{%
\begin{tabular}{ccccc}
\hline
\begin{tabular}[c]{@{}c@{}}Number of Nodes\end{tabular} &
  \begin{tabular}[c]{@{}c@{}}FLOPs (Millions)\\ of Teacher model\end{tabular} &
  \begin{tabular}[c]{@{}c@{}}model size (MB)\\ of Teacher model\end{tabular} &
  \begin{tabular}[c]{@{}c@{}}Teacher Model Accuracy (\%)\end{tabular} &
  \begin{tabular}[c]{@{}c@{}}Student Model Accuracy (\%)\end{tabular} \\ \hline
32  & 34.54  & 0.54 & 98.83 & 97.25 \\
64  & 44.37  & 0.69 & 99.06 & 97.34 \\
128 & 59.05  & 0.92 & 99.82 & 97.76 \\
256 & 126.16 & 1.97 & 99.67 & 97.58 \\ \hline
\end{tabular}%
}
\end{table}

\subsection{Evaluating the Superiority of ELMMSD for Subdomain Alignment}

To evaluate the effectiveness of deep learning models in aligning source and target domain distributions, the $\mathcal{A}$-distance metric \cite{ben2006analysis,ben2010theory} was employed. This metric is approximated by training a linear SVM classifier to distinguish between samples from the source and target domains, defined as:  
\begin{equation}
\hat{d}_{A} = 2(1-2\zeta),
\end{equation}  
where $\zeta$ represents the classification error. While $\mathcal{A}$-distance measures global domain adaptation, it does not account for subdomain alignment. To address this limitation, the $\mathcal{A}_{L}$-distance metric \cite{zhu2020deep} was adopted to quantify subdomain discrepancies:  
\begin{equation}
d_{{\mathcal{A}}_{L}} = 2 \sum_{c=1}^{C} p(c)\left(1-2\zeta^{c}\right),
\end{equation}  
where $\zeta^{c}$ denotes the classifier error for each health category under varying loads, and $p(c)$ is the probability of class $c$ in the target domain.

\begin{figure}
    \centering
    \includegraphics[width=\columnwidth]{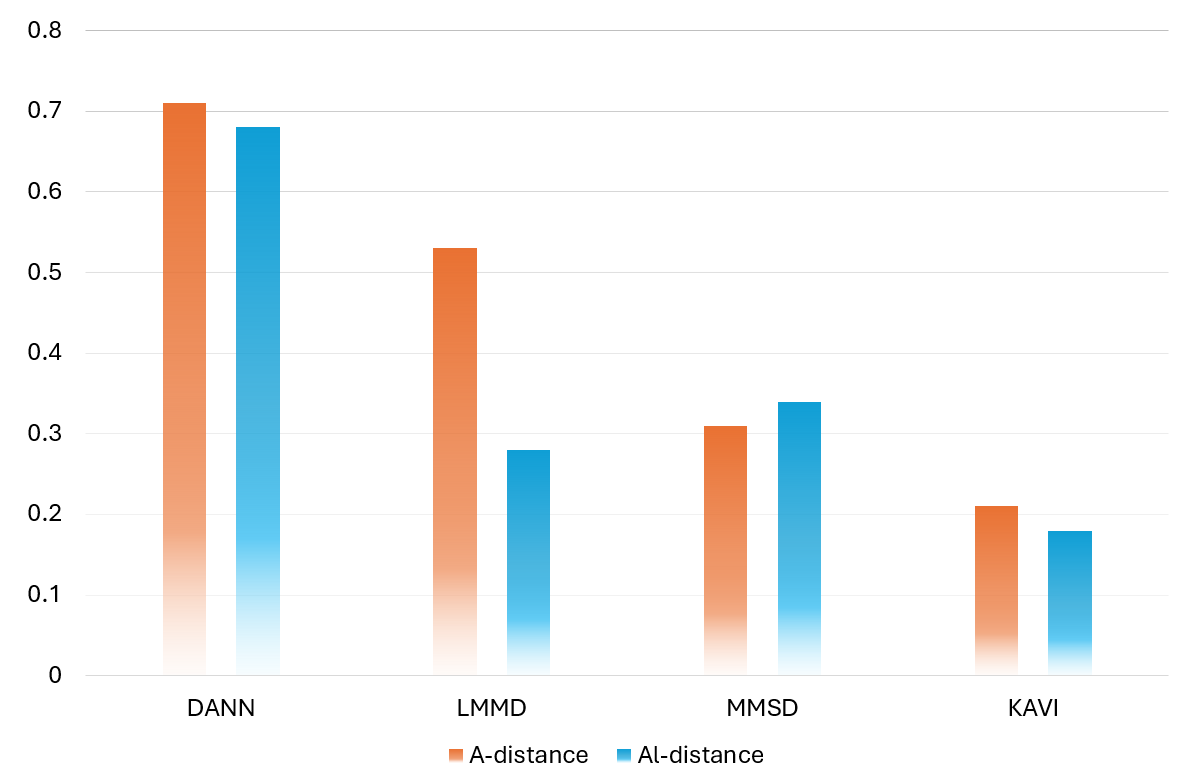}
    \caption{Comparison $\mathcal{A}-\text{distance}$ and $\mathcal{A}_{L}-\text{distance}$ for various SDA and DA methods on task J3 $\to$ J1}
    \label{a-distance}
\end{figure}

Both $\mathcal{A}$-distance and $\mathcal{A}_{L}$-distance were calculated for various DA and SDA methods to quantify their effectiveness in reducing domain discrepancies. As illustrated in Fig. \ref{a-distance} for task J3$\to$J1, the proposed KAVI method with ELMMSD significantly reduced both global and subdomain discrepancies compared to competing approaches, including LMMD, DANN, and MMSD. These results demonstrate the superior capability of KAVI with ELMMSD in mitigating global distribution gaps and achieving precise subdomain alignment.

\subsection{Analyzing the effect of label smoothing on Teacher model performance
}

To assess the impact of label smoothing on the Teacher model within the proposed KAVI framework, we conducted an ablation study by comparing the model's accuracy with and without label smoothing. Table~\ref{label_smoothing} summarizes the results across multiple transfer tasks. The incorporation of label smoothing consistently enhances accuracy across all tasks. For instance, accuracy for A1 $\to$ A4 improves from 99.18\% to 99.67\%, while for J3 $\to$ J2, it increases from 98.80\% to 99.09\%. These results demonstrate that label smoothing effectively reduces model overconfidence, fostering better generalization and improved parameter space exploration. This enhancement enables more accurate alignment of marginal and conditional distributions during subdomain adaptation, underscoring the critical role of label smoothing in achieving superior performance in fault diagnosis tasks.

% \usepackage{graphicx}
\begin{table}[]
\caption{Comparison of Teacher model accuracy: with and without label smoothing}
\label{label_smoothing}
\resizebox{\columnwidth}{!}{%
\begin{tabular}{ccccccc}
\hline
Method & A1 $\to$ A4 & A3 $\to$ A2 & A4 $\to$ A2 & J1 $\to$ J2 & J1 $\to$ J3 & J3 $\to$ J2 \\ \hline
KAVI (hard label)  & 99.18       & 99.59       & 98.33       & 98.24       & 98.63       & 98.80       \\
KAVI (label smoothing) & 99.67       & 99.83       & 98.90       & 98.93       & 98.87       & 99.09       \\ \hline
\end{tabular}%
}
\end{table}

% \subsection{highlights}
% \begin{itemize}
%     \item Resource-constrained student model with knowledge distillation for fault diagnosis.
%     \item GCN with ARMA filters enhances feature extraction in subdomain adaptation.
%     \item ELMMSD improves subdomain alignment using mean and variance statistics in RKHS.
%     \item Smooth labeling increases subdomain distance, enhancing diagnostic accuracy.
%     \item Teacher-student model improves efficiency while maintaining diagnostic performance.
% \end{itemize}

\section{conclusion}
This paper proposes a novel approach for bearing fault diagnosis under varying working conditions, focusing on resource-constrained environments. We introduce a progressive knowledge distillation framework that transfers knowledge from a complex teacher model (including GCN with ARMA filters) to a lightweight student model, ensuring efficient deployment. The ELMMSD method improves subdomain alignment by leveraging both mean and variance statistics in the RKHS, enhancing the reliability of subdomain discrepancy calculations. We further demonstrate the effectiveness of the ARMA filter through comparison with state-of-the-art filters, highlighting its superior performance in subdomain adaptation by extracting better geometric features from the data. The use of smooth labeling techniques is shown to increase accuracy by improving the distance between subdomains. Additionally, combining knowledge distillation with subdomain adaptation demonstrates improved fault diagnosis performance. While this study assumes identical sample labels between source and target domains, the proposed method remains effective for practical applications. 
\par Future research will explore addressing open-set fault diagnosis, further enhancing the generalizability and applicability of this method for intelligent fault diagnosis in diverse devices and environments.

% \vspace{1em}

% \noindent \textbf{CRediT authorship contribution statement:} \\
% \textbf{Mohammadreza Kavianpour:} Writing- Original draft, Conceptualization, Methodology, Software, Supervision preparation, Project administration. \\
% \textbf{Parisa Kavianpour:} Software, Resources, Methodology, Visualization, Writing - Review \& Editing. \\
% \textbf{Amin Ramezani:} Validation, Formal analysis, Resources. \\
% \textbf{Mohammad TH Behesht:} Formal analysis, Data Curation, Validation. \\

% \vspace{0.25em}

% \noindent \textbf{Declaration of Competing Interest:} \\
% The authors declare that they have no known competing financial interests or personal relationships that could have appeared to influence the work reported in this paper. \\

% \vspace{0.25em}

% \noindent \textbf{Data Availability:} \\
% The datasets generated during and/or analyzed during the current study are available from the corresponding author upon reasonable request.

\bibliographystyle{ieeetr}  % Choose the style according to your preference
\EOD

\end{document}